# Technical Report on BaumEvA Evolutionary Optimization Python-Library Testing

Vadim Tynchenko [0000-0002-3959-2969], Aleksei Kudryavtsev [0009-0000-2060-992X],
Vladimir Nelyub [0000-0003-4263-2367], Aleksei Borodulin [0000-0002-9648-2395],
Andrei Gantimurov

Bauman Moscow State Technical University, Moscow, Russia
vadimond@mail.ru

**Abstract.** This report presents the test results Python library BaumEvA, which implements evolutionary algorithms for optimizing various types of problems, including computer vision tasks accompanied by the search for optimal model architectures. Testing was carried out to evaluate the effectiveness and reliability of the proposed methods, as well as to determine their applicability in various fields. During testing, various test functions and parameters of evolutionary algorithms were used, which made it possible to evaluate their performance in a wide range of conditions. Test results showed that the library provides effective and reliable methods for solving optimization problems. However, some limitations were identified related to computational resources and execution time of algorithms on problems with large dimensions. The report includes a detailed description of the tests performed, the results obtained and conclusions about the applicability of the genetic algorithm in various tasks. Recommendations for choosing algorithm parameters and using the library to achieve the best results are also provided. The report may be useful to developers involved in the optimization of complex computing systems, as well as to researchers studying the possibilities of using evolutionary algorithms in various fields of science and technology.

## 1    Introduction

BaumEvA (version 0.6.3) – a library on programming language Python (supported versions 3.7, 3.8, 3.9, 3.10, 3.11), on evolutionary algorithms is a set of tools and methods used to solve optimization problems [1]. Evolutionary algorithms (EA) are a type of metaheuristic algorithms that are based on simulating natural processes such as natural selection, genetic mutations and crossovers.

The main components of evolutionary algorithms include population, genetic operator, fitness function and selection process. The population is the possible solutions to the optimization problem, which are evaluated using the fitness function. Genetic operators such as crossover and mutation are used to generate new solutions from existing ones. The selection process determines which solutions will be selected for the next iteration.

BaumEvA provides the following features:

1. Optimization tasks:
   - Unconditional optimization
   - Conditional optimization
   - Multi-criteria optimization (version 0.7.0)

2. Different types of evolutionary algorithms:

   - Binary genetic algorithm
   - Categorical genetic algorithm
   - Combinatorial genetic algorithm

3. Types of populations:

   - Binary population
   - Binary population with Gray codes
   - Categorical population
   - Ordered categorical population – combinatorial

4. Fitness functions:

   - Hyperbolic $fitness = \frac{1}{1-|f(x)-optimum|}$

5. Selections:

   - Tournament
   - Ranked
   - Roulette Wheel

6. Crossovers:

   - One-point
   - Two-point
   - Uniform
   - Order crossover (OX1)

7. Mutation

   - Classic binary mutation
   - Categorical mutation
   - Inversion mutation
   - Movement mutation

- Swap mutation
- Shift mutation

8. Elitism is the transfer of a certain number of the best or random individuals from the previous generation.

9. Penalty functions:

   - Static penalties (version 0.6.4)
   - Dynamic penalties
   - Adaptive penalties (version 0.6.4)

The library allows the user to flexibly configure the parameters of the algorithm to solve the optimization problem. On the user interface side, several levels of abstraction are implemented:

1. Ease mode – use of ready-made GA implementations (BinaryGA, CombinatoryGA, CategoricalGA). Easy to use, but have limited capabilities.
2. Collector mode – use of the CollectorGA class, which allows your own GA using operators and library methods. More flexible to use than Easy mode, but requires a deeper understanding of evolutionary algorithms and knowledge of library tools.
3. Advanced mode – interaction directly with genetic algorithm operators. It has almost unlimited flexibility, the ability to write iterations yourself, but at the same time it requires a good level of understanding of the GA and all the features of the library.

## 2       Materials and Methods

When testing the library, both conditional and unconditional optimization problems were used. The purpose of testing was to check the performance of the implemented genetic algorithm. The performance of the algorithm was assessed as achieving the optimum with a given accuracy in a limited number of calculations of the objective function or as the number of calculations of the objective function required to achieve the optimum. For each type of problem, the most suitable GA operators and parameters were selected.

### A.  Conditional optimization

Problems for conditional optimization obtained from competition CEC2017 [2]. Formulation of the problem:

Minimize: $f(X), X = (x_1, x_2, \ldots, x_n)$ $and$ $X \in S$.

Subject to: $g_i(X) \leq 0, i = 1, \ldots, p; \ h_j(X) = 0, j = p + 1, \ldots, m$

Usually, equality constraints are transformed into inequalities of the form:

$$|h_i(X)| - \varepsilon \leq 0, for \ j = p + 1, \ldots, m, \qquad \varepsilon = 0.0001$$

Task CO1

$$Min\ f(x) = \sum_{i=1}^{D} \left( \sum_{j=1}^{i} z_j \right)^2, z = x - shift$$

$$g(x) = \sum_{i=1}^{D} [z_i^2 - 5000 \cos(0.1\pi z_i) - 4000] \leq 0$$

$$x \in [-100,100]^D$$

This problem was tested with dimensions D10, D30, D50 and D100. For each dimension, 3 levels of restrictions on the calculation of the objective function (MaxFEs) were allocated: 2*10e+4, 10e+5, 2*10e+5. For each constraint, 30 independent GA runs were available, from which the best, worst, average and median solution were selected (Table I-III).

GA parameters:

- Number of individuals 25;
- The number of generations depends on the maximum number of available calculations of the objective function and varied from 800 to 8000;
- Grid step 0.0001;
- Binary population with Gray codes;
- Ranked selection;
- Two-point crossover;
- Strong mutation;
- Elitism (top 5%).

### B. Unconditional optimization

Problems for unconditional optimization obtained from competition CEC2017 [3]. The problem statement is similar to conditional optimization: $Min\ F(x), x \in [-100, 100]^D, D = \{10, 30, 50, 100\}$, MaxFEs (maximum available number of function calculations) = 10000*D.

1. Bent Cigar Function

$$f(x) = x_1^2 + 10^6 \sum_{i=2}^{D} x_i^2$$

2. Zakharov Function

$$f(x) = \sum_{i=1}^{D} x_i^2 + \left( \sum_{i=1}^{D} 0.5x_i \right)^2 + \left( \sum_{i=1}^{D} 0.5x_i \right)^4$$

3. Rosenbrock's Function

$$f(x) = \sum_{i=1}^{D-1} (100(x_i^2 - x_{i+1})^2 + (x_i - 1)^2)$$

4.  Rastrigin's Function

$$f(x) = \sum_{i=1}^{D} (x_i^2 - 10 \cos(2\pi x_i) + 10)$$

5.  Expanded Schaffer Function

$$f(x) = g(x_1, x_2) + g(x_2, x_3) + \cdots + g(x_{D-1}, x_d) + g(x_D, x_1)$$

$$g(x, y) = 0.5 + \frac{\left(sin^2\left(\sqrt{x^2 + y^2}\right) - 0.5\right)}{(1 + 0.001(x^2 + y^2))^2}$$

6.  Levy Function

$$f(x) = sin^2(\pi w_1)$$
$$+ \sum_{i=1}^{D-1} (w_i - 1)^2 [1 + 10 sin^2(\pi w_i + 1)]$$
$$+ (w_D - 1)^2 [1 + sin^2(2\pi w_D)]$$

$$w_i = 1 + \frac{x_i - 1}{4}, \forall i = 1, \dots D$$

7.  High Conditioned Elliptic Function

$$f(x) = \sum_{i=1}^{D} (10^6)^{\frac{i-1}{D-1}} x_i^2$$

8.  Discus Function

$$f(x) = 10^6 x_1^2 + \sum_{i=2}^{D} x_i^2$$

9.  Ackley's Function

$$f(x) = -20 exp\left(-0.2 \sqrt{\frac{1}{D}\sum_{i=1}^{D} x_i^2}\right) - exp\left(\frac{1}{D}\sum_{i=1}^{D} \cos(2\pi x_i)\right) + 20 + e$$

10. Weierstrass Function

$$f(x) = \sum_{i=1}^{D} \left( \sum_{k=0}^{k_{max}} [a^k \cos(2\pi b^k (x_i + 0.5))] \right) - D \sum_{k=0}^{k_{max}} [a^k \cos(\pi b^k)]$$

$$a = 0.5, b = 3, k_{max} = 20$$

11. Griewank's Function

$$f(x) = \sum_{i=1}^{D} \frac{x_i^2}{4000} - \prod_{i=1}^{D} \cos\left(\frac{x_i}{\sqrt{i}}\right) + 1$$

12. Katsuura Function

$$f(x) = \frac{10}{D^2} \prod_{i=1}^{D} \left( 1 + i \sum_{j=1}^{32} \frac{2^j x_i - round(2^j x_i)}{2^j} \right)^{\frac{10}{D^{12}}} - \frac{10}{D^2}$$

13. HappyCat Function

$$f(x) = \left| \sum_{i=1}^{D} x_i^2 - D \right|^{\frac{1}{4}} + \frac{0.5 \sum_{i=1}^{D} x_i^2 + \sum_{i=1}^{D} x_i}{D} + 0.5$$

14. HGBat Function

$$f(x) = \left| \left( \sum_{i=1}^{D} x_i^2 \right)^2 - \left( \sum_{i=1}^{D} x_i \right)^2 \right|^{\frac{1}{2}} + \frac{0.5 \sum_{i=1}^{D} x_i^2 + \sum_{i=1}^{D} x_i}{D} + 0.5$$

Basic parameters of GA:

- Number of individuals 25;
- The number of generations depends on the maximum number of available calculations of the objective function and varied from 4000 to 40000;
- Grid step 10e-8;
- Binary population with Gray codes;
- Tournament selection, tournament size is 3;
- One-point crossover;
- Strong mutation;
- Elitism (top 5%).

GA parameters may vary depending on the task. The number of runs of the algorithm is limited to 51. Every 1% of MaxFEs the error was calculated (the difference between the minimum function and the best solution). If the error was less than 10e-8, then it was equal to 0 (this rule was a criterion for stopping the algorithm, along with MaxFEs). For each task, a graph of the error depending on the used computing resources is given for the dimension D10 (51 runs). For other dimensions, such a graph is not available due to the very large time costs (for D50, one GA launch can take up to 25 minutes). There is also a general Table IV for each problem with the following parameters: best solution, worst, average, median, standard deviation among all GA runs (51).

### C. Binary tasks

Below are the various Boolean functions [4]. To test the GA, the dimensions D50, D100, D200, D500, D1000 were used. For each dimension, the GA was run 51 times. There was no limit on the number of function evaluations (FEs), the algorithm worked until it found the desired solution. For each dimension, a graph of the error versus FEs is given for each run, and the average number of FEs required for one run, as well as the minimum, maximum and standard deviation in Table V.

#### 1. OneMax Function

A trivial Boolean problem: you need to maximize the number of 1's in a given vector. This problem is a classic example of a linear problem in which there is no connection between the vector values.

$$f(\langle x_1, \dots, x_n \rangle) = \sum_{i=1}^{n} x_i$$

#### 2. LeadingOnes Function

It is necessary to maximize the number of consecutive unit digits, if counted from the beginning of the vector. In other words, the position of the first zero bit in the vector plays the most important role. The "Leading Ones" problem is nonlinear: the contribution ("utility") of component xi significantly depends on the values of components $x_1$, ..., $x_{i-1}$.

$$f(\langle x_1, \dots, x_n \rangle) = \sum_{i=1}^{n} \prod_{j=1}^{i} x_j$$

#### 3. Trap function

So-called decoy problems are a classic example of deceptive functions. The fitness of a vector is equal to the number of zero digits if the vector contains at least one unit digit, but if all the digits are unit ones, then the fitness suddenly increases to the value n+1 (n is the length of the vector). Thus, in this problem the search direction is set, leading away from the optimal solution ("no zero digits"), straight into the "trap". For example, *f (<0, 0, 0, 0>) = 4*, *f (<0, 0, 1, 0>) = 3*, *f (<1, 0, 0, 1>) = 2*, *f (<1, 0, 1, 1>) = 1*, but *f (<1, 1, 1, 1>) = 5*. The mathematical formulation of the problem includes two terms: the first is equal to the number of zero digits, and the second appears only when all digits are single.

$$f(\langle x_1, \ldots, x_n \rangle) = \left( n - \sum_{i=1}^{n} x_i \right) + (n+1) \prod_{i=1}^{n} x_i$$

GA parameters:

- Number of individuals 10;
- Number of generations – until we find a solution;
- Boolean grid step (1);
- Binary population;
- Tournament selection, tournament size is 3;
- Uniform crossover;
- Normal mutation;
- Elitism (top 5%).

## 3    Results

### A.    Conditional optimization

Below are the results of conditional optimization in Table 1 and Fig. 1-4. The graphs present the results of the best solutions obtained after each GA run for various restrictions on the maximum number of calculations.

**Table 1.** Solutions for each dimension and MaxFEs for conditional optimization function CO1.

| Dimension | Solutions/ FEs | Best Solution | Worst Solution | Mean Solution | Median Solution |
|:---:|:---:|:---:|:---:|:---:|:---:|
| **D10** | 2*10e+4 | 0.0596 | 8.1426 | 2.1010 | 1.5298 |
| | 10e+5 | 0.0003 | 0.0148 | 0.0034 | 0.0018 |
| | 2*10e+5 | 0* | 0.0007 | 0.0002 | 0.0001 |
| **D30** | 2*10e+4 | 2043.3 | 7528.3 | 4371.6 | 4180.6 |
| | 10e+5 | 19.5 | 221.9 | 100.7 | 79.9 |
| | 2*10e+5 | 2.94 | 16.13 | 9.17 | 8.59 |
| **D50** | 2*10e+4 | 12268.1 | 33272.8 | 21801.7 | 21291.1 |
| | 10e+5 | 1300.6 | 4847.1 | 2895.8 | 2755.5 |
| | 2*10e+5 | 269.4 | 966.8 | 569.7 | 538.1 |
| **D100** | 2*10e+4 | 85773.2 | 139010.9 | 111074.7 | 109628.9 |
| | 10e+5 | 33902.9 | 53685.6 | 40846.2 | 39174.7 |
| | 2*10e+5 | 13643.9 | 23894.0 | 18819.1 | 19134.8 |

* - solution with an accuracy of less than 0.0001 is equal to 0.

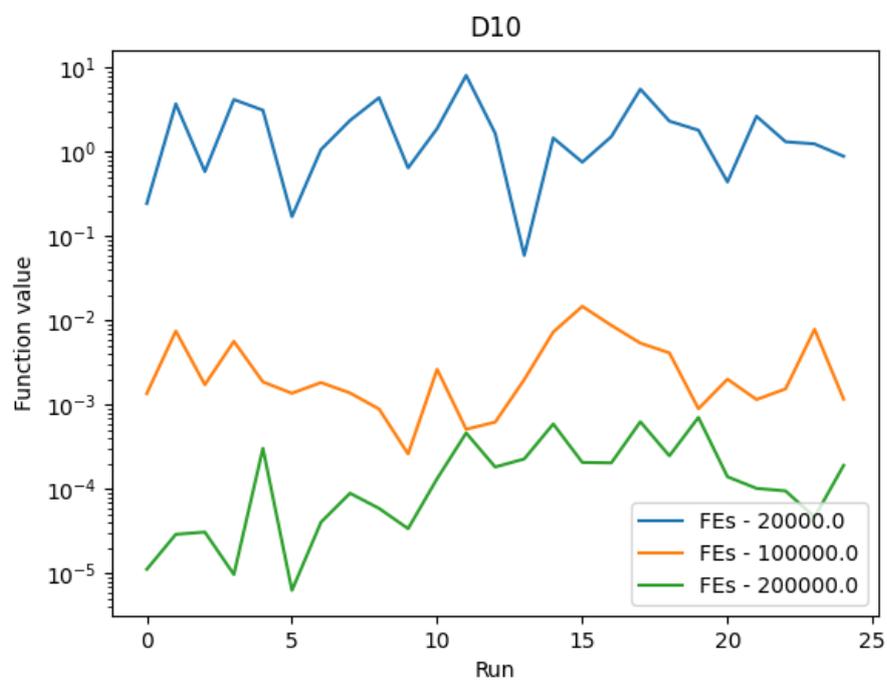

**Fig. 1.** Best function value by each run with dimension 10.

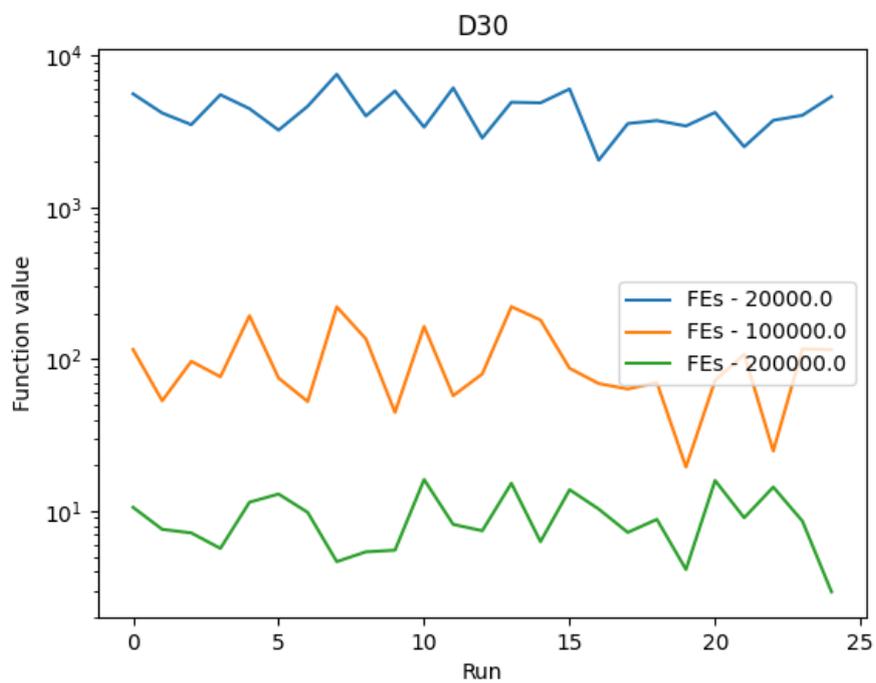

**Fig. 2.** Best function value by each run with dimension 30.

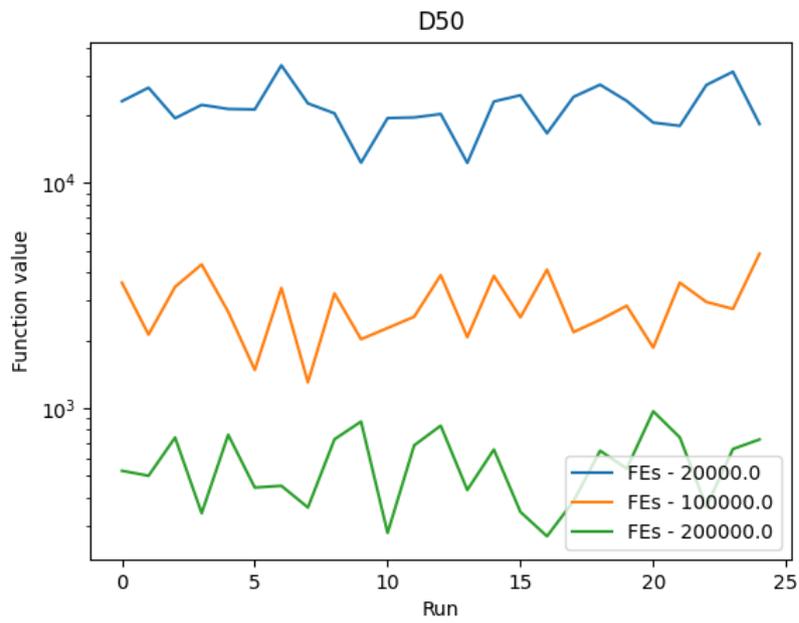

**Fig. 3.** Best function value by each run with dimension 50.

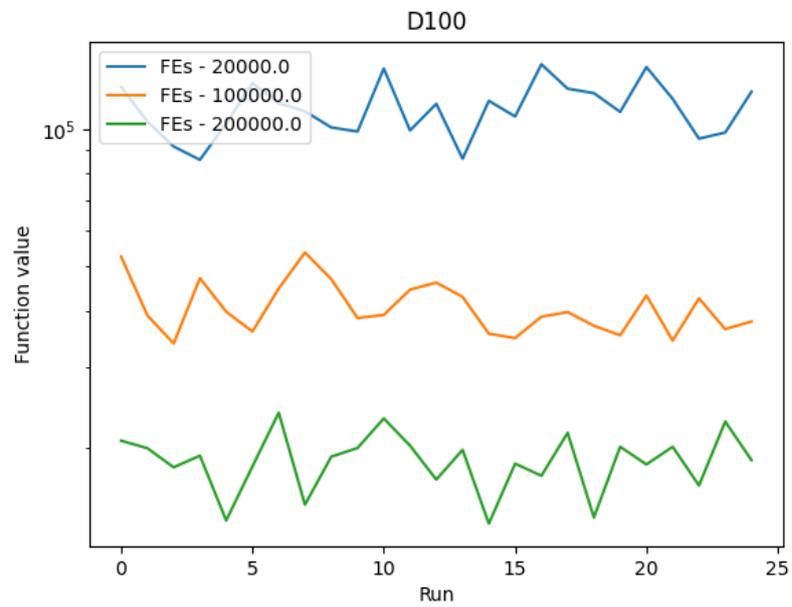

**Fig. 4.** Best function value by each run with dimension 100.

The minimum of a function with a given accuracy (0.0001) and restrictions on the number of calculations was found only for dimension D10. When the grid step is reduced by 10 times (to 0.00001), the results improve, which is shown in the Table 2 in bold for the D10 dimension and in the Fig. 5 below.

**Table 2.** Solutions for D10 and 10e-5 grid step for conditional optimization function CO1.

| Dimension | Solutions/FEs | Best Solution | Worst Solution | Mean Solution | Median Solution |
|-----------|---------------|---------------|----------------|---------------|-----------------|
| **D10** | 2*10e+4 | 0.0653 | 8.8238 | **1.8120** | **0.9052** |
| | 10e+5 | 0.0003 | **0.0064** | **0.0021** | **0.0014** |
| | 2*10e+5 | **0\*** | **0.0005** | **0\*** | **0\*** |

\* - solution with an accuracy of less than 0.0001 is equal to 0.

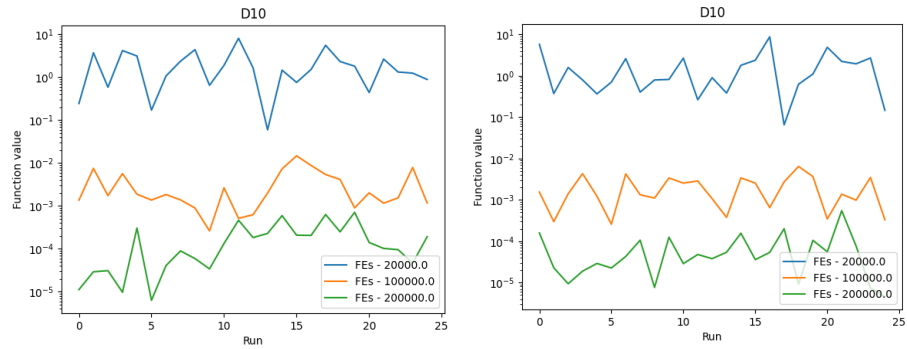

**Fig. 5.** Best function value by each run with dimension 10 and the grid step is 10e-4 on the left and 10e-5 on the right.

When the grid step is increased by 10 times (up to 10e-3), the results deteriorate, as shown in the Table 3 and Fig. 6. below (D10 dimension).

**Table 3.** Solutions for D10 and 10e-3 grid step for conditional optimization function CO1.

| Dimension | Solutions/FEs | Best Solution | Worst Solution | Mean Solution | Median Solution |
|-----------|---------------|---------------|----------------|---------------|-----------------|
| **D10** | 2*10e+4 | 0.2243 | 11.4168 | 2.3373 | 1.5403 |
| | 10e+5 | 0.0014 | 0.0408 | 0.0096 | 0.0072 |
| | 2*10e+5 | 0\* | 0.0027 | 0.0007 | 0.0007 |

\* - solution with an accuracy of less than 0.0001 is equal to 0.

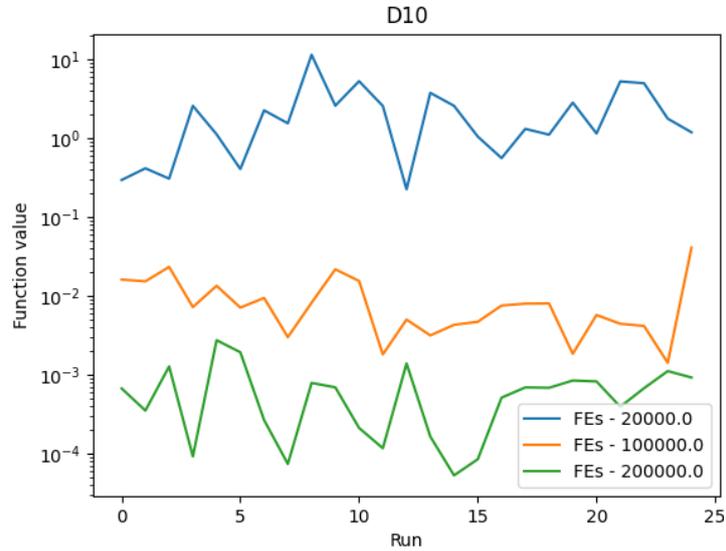

**Fig. 6.** Best function value by each run with dimension 10 and the grid step is 10e-3.

Fig. 7 demonstrates the dynamics of changes in the values of the fitness function of the best, average and worst individual in each generation for dimension D100, without restrictions on the number of calculations. The value of the fitness function tends to 1, which corresponds to the objective function tending to 0 (minimum).

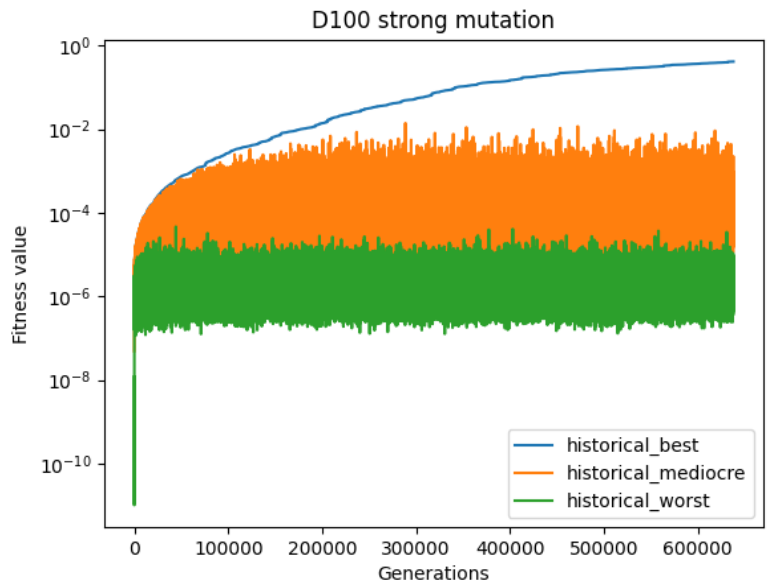

**Fig. 7.** Fitness function of best, mediocre and worst individuals in each generation on dimensional 100 and strong mutation.

### B. *Unconditional optimization*

Below are the results of unconstrained optimization. Fig. 8-21 represent the change in error (the difference in absolute value between the optimum of the function and the current best solution of the algorithm) depending on the amount of computing resources spent as a percentage for each of the 51 runs of the GA. Table 4 presents the results of the best solution, worst solution, average, median and standard deviation among all (51) GA runs for each of the 14 functions.

**Table 4.** Solutions for D10 and 10e-8 grid step for unconditional optimization function.

| Type func. | Number func. | Best | Worst | Mean | Median | Std |
|---|---|---|---|---|---|---|
| **Unimodal** | 1 | 0* | 0* | 0* | 0* | 0* |
| | 2 | 0* | 0* | 0* | 0* | 0* |
| **Multimodal** | 3 | 0.0196 | 85.6854 | 35.4461 | 39.0821 | 27.2956 |
| | 4 | 2.9849 | 9.9496 | 6.0673 | 5.9698 | 1.4778 |
| | 5 | 0* | 1.2357 | 0.4352 | 0.3173 | 0.3474 |
| | 6 | 0* | 0.4543 | 0.0089 | 0* | 0.0630 |
| **Unimodal** | 7 | 0* | 0* | 0* | 0* | 0* |
| | 8 | 0* | 0* | 0* | 0* | 0* |
| **Multimodal** | 9 | 0* | 1.04e-5 | 5.10e-7 | 2.4e-7 | 1.4e-6 |
| | 10 | 8.68e-5 | 0.0583 | 0.0029 | 0.0006 | 0.0084 |
| | 11 | 0.0123 | 0.5467 | 0.1342 | 0.1033 | 0.1294 |
| | 12 | 0* | 1.36e-7 | 0* | 0* | 0* |
| | 13 | 0.0319 | 0.2566 | 0.1391 | 0.1396 | 0.0424 |
| | 14 | 0.0554 | 0.8232 | 0.2698 | 0.2359 | 0.1549 |

\* - solution with an accuracy of less than 10e-8 is equal to 0.

### 1. Bent Cigar Function

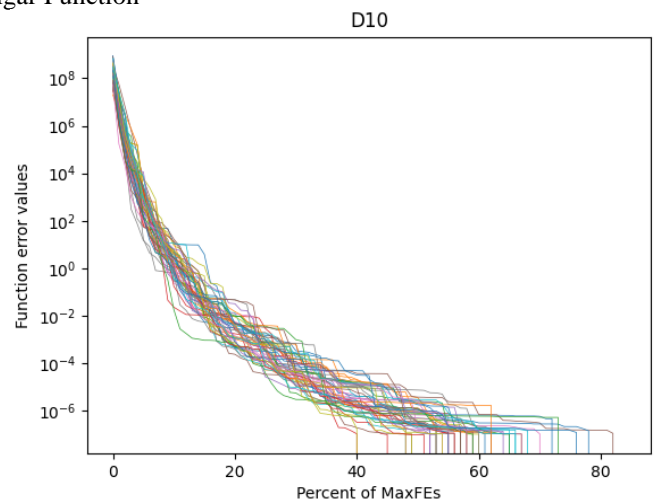

**Fig. 8.** Percentage of MaxFEs (for each of 51 runs) usage to solve the Bent Cigar function with a given accuracy of 10e-8.

2. Zakharov Function

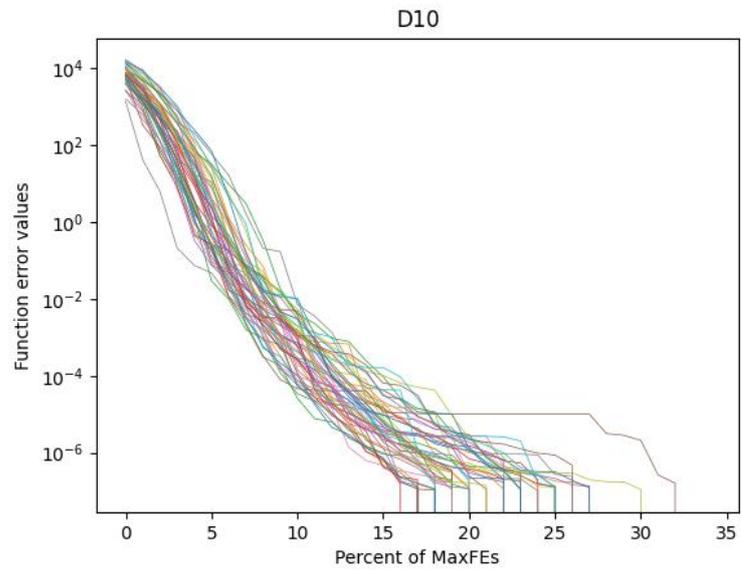

**Fig. 9.** Percentage of MaxFEs (for each of 51 runs) usage to solve the Zakharov function with a given accuracy of 10e-8.

3. Rosenbrock's Function

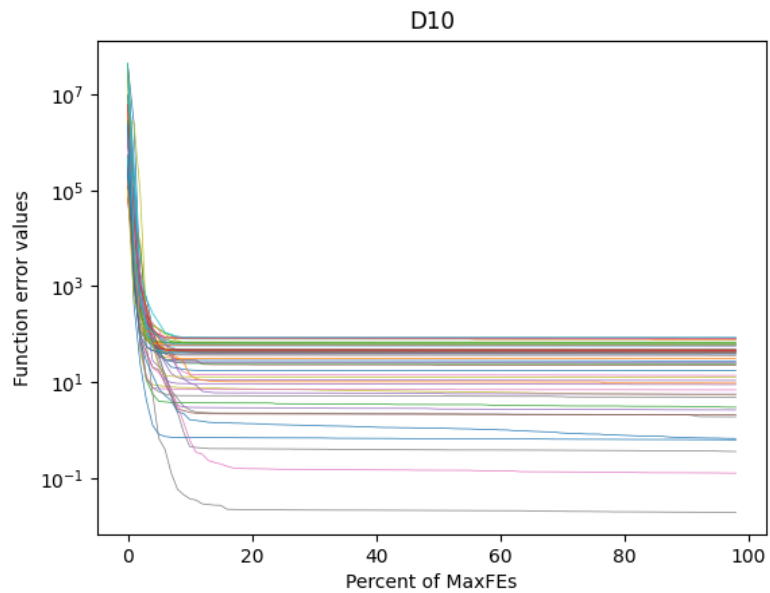

**Fig. 10.** Percentage of MaxFEs (for each of 51 runs) usage to solve the Rosenbrock's function with a given accuracy of 10e-8.

4.   Rastrigin's Function

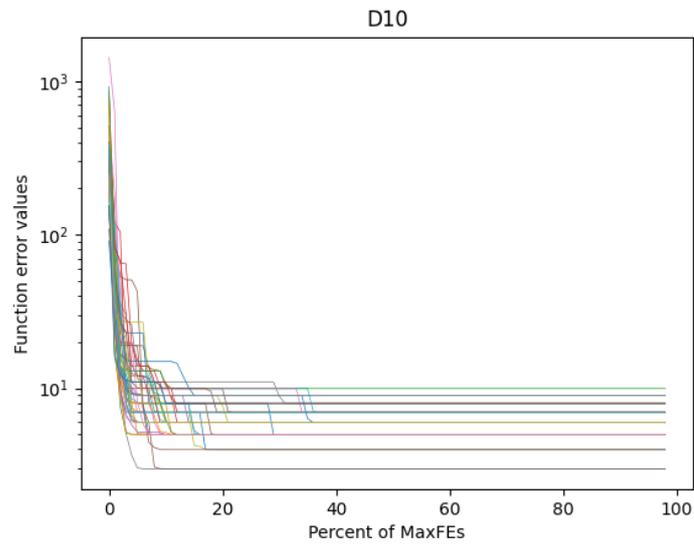

**Fig. 11.** Percentage of MaxFEs (for each of 51 runs) usage to solve the Rastrigin's function with a given accuracy of 10e-8.

5.   Expanded Schaffer Function

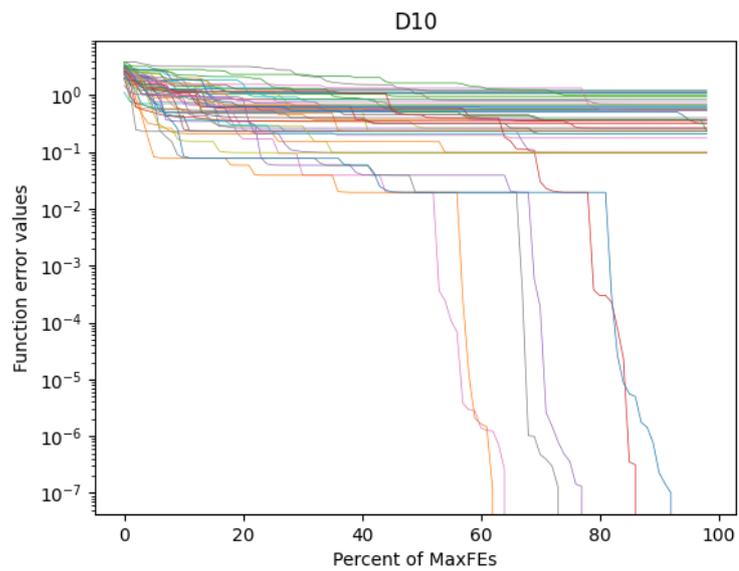

**Fig. 12.** Percentage of MaxFEs (for each of 51 runs) usage to solve the Expanded Schaffer function with a given accuracy of 10e-8.

6.  Levy Function

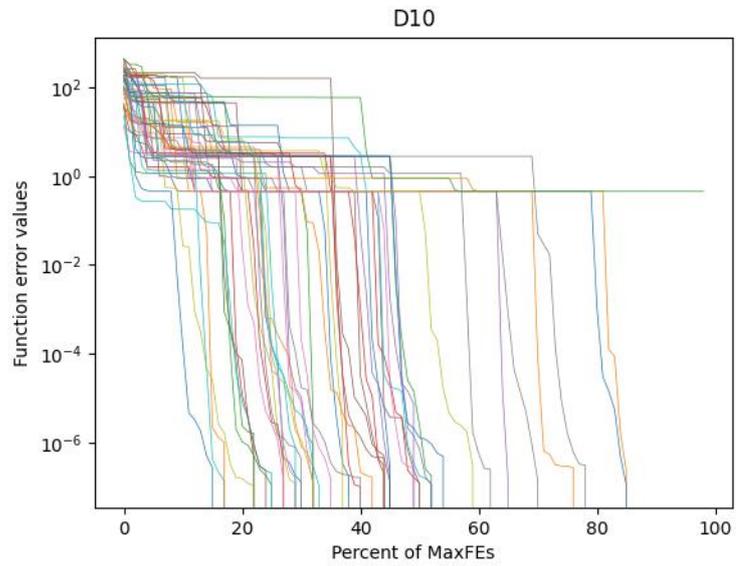

**Fig. 13.** Percentage of MaxFEs (for each of 51 runs) usage to solve the Levy function with a given accuracy of 10e-8.

7.  High Conditional Elliptic Function

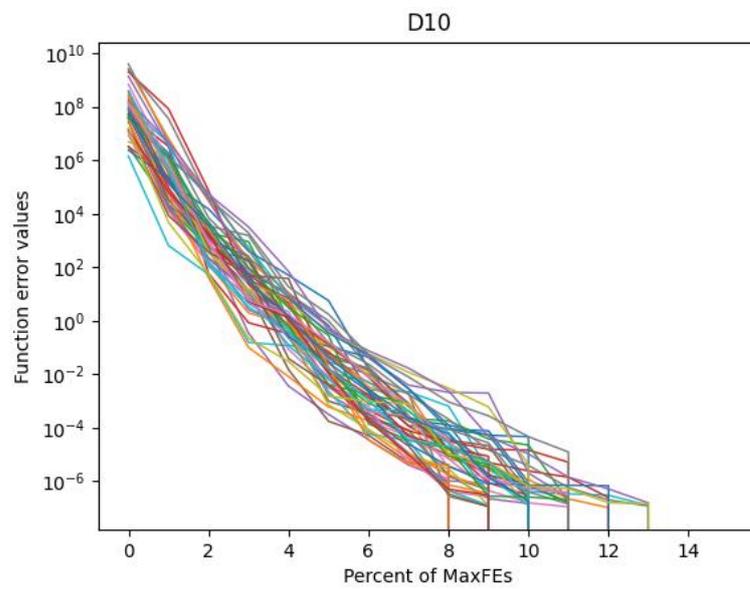

**Fig. 14.** Percentage of MaxFEs (for each of 51 runs) usage to solve the High Conditioned Elliptic function with a given accuracy of 10e-8.

8. Discus Function

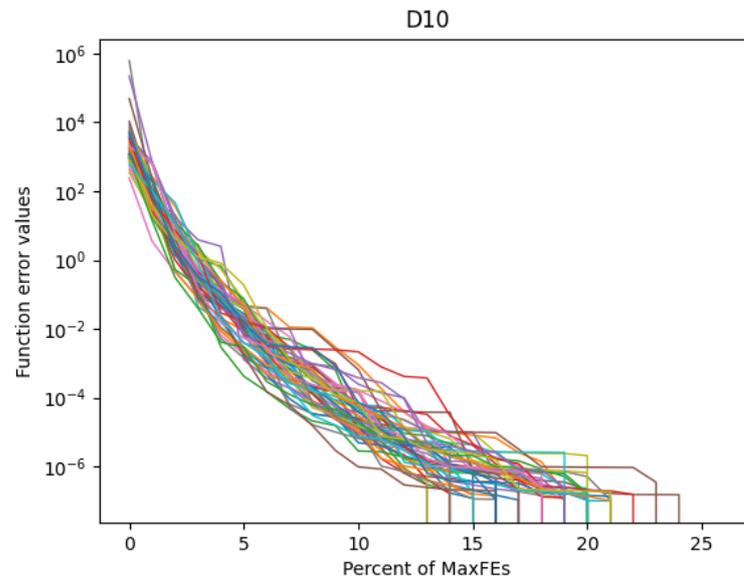

**Fig. 15.** Percentage of MaxFEs (for each of 51 runs) usage to solve the Discus function with a given accuracy of 10e-8.

9. Ackley's Function

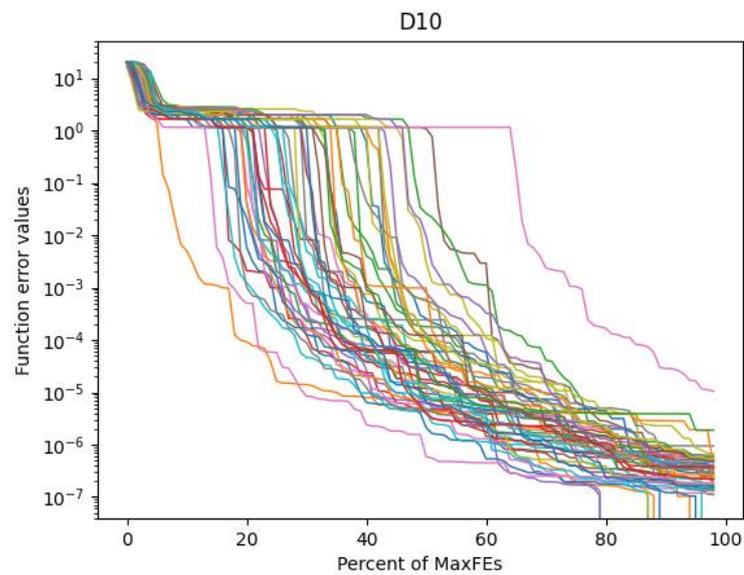

**Fig. 16.** Percentage of MaxFEs (for each of 51 runs) usage to solve the Ackley's function with a given accuracy of 10e-8.

## 10. Weierstrass Function

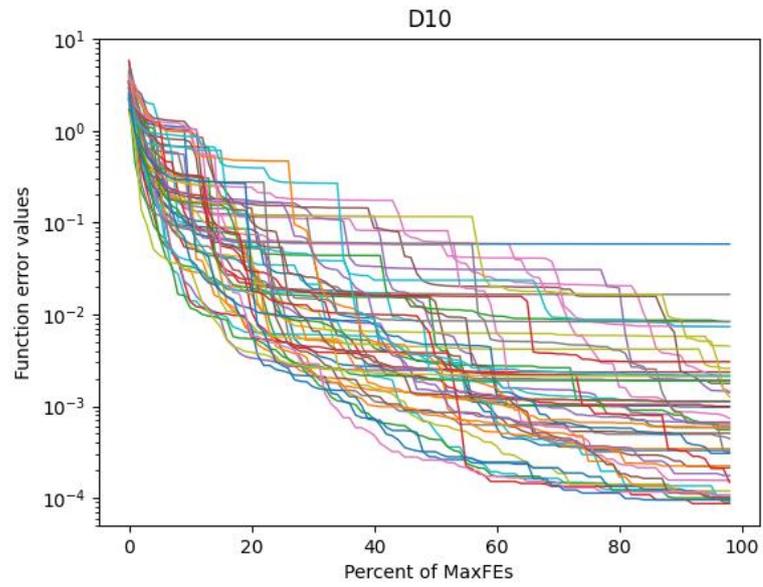

**Fig. 17.** Percentage of MaxFEs (for each of 51 runs) usage to solve the Weierstrass function with a given accuracy of 10e-8.

## 11. Griewank's Function

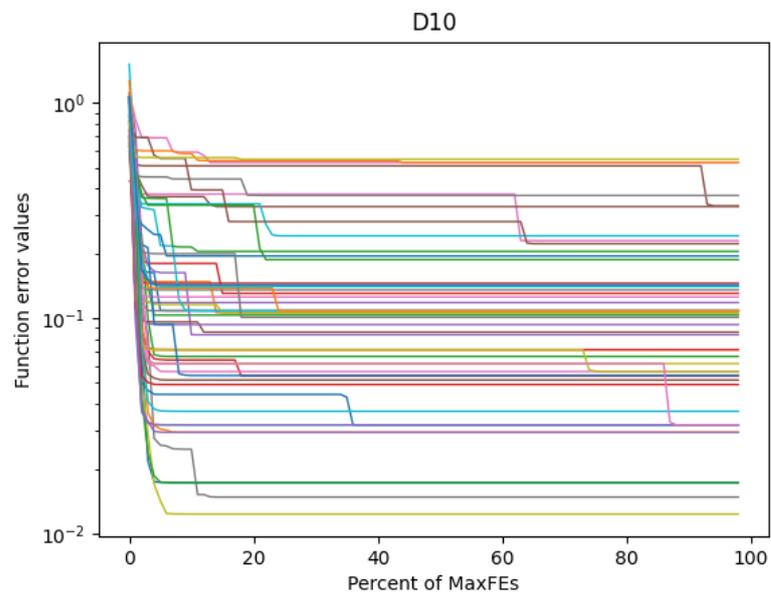

**Fig. 18.** Percentage of MaxFEs (for each of 51 runs) usage to solve the Griewank's function with a given accuracy of 10e-8.

12. Katsuura Function

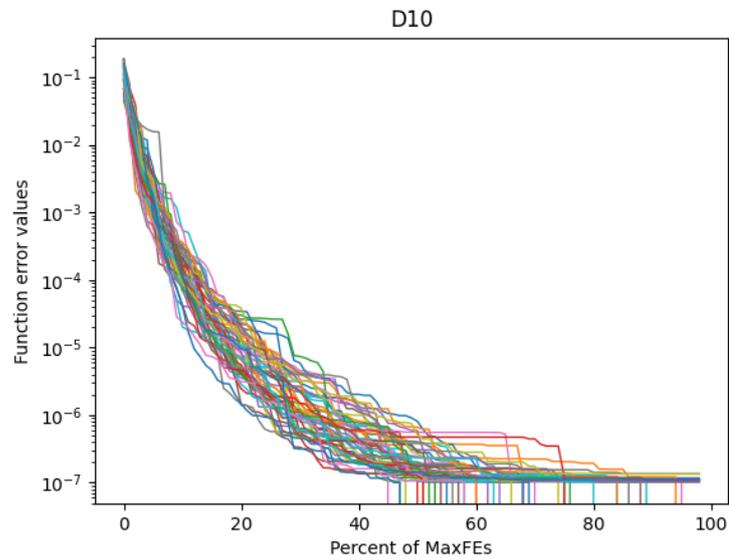

**Fig. 19.** Percentage of MaxFEs (for each of 51 runs) usage to solve the Katsuura function with a given accuracy of 10e-8.

13. HappyCat Function

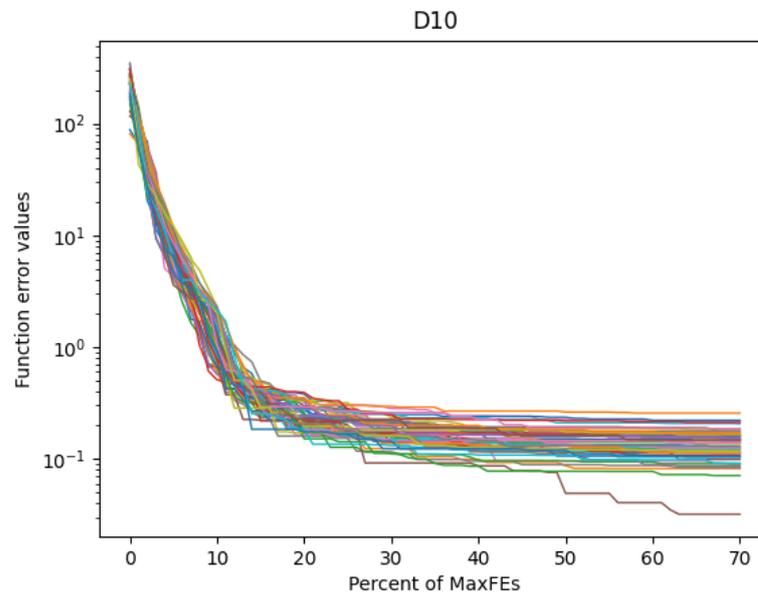

**Fig. 20.** Percentage of MaxFEs (for each of 51 runs) usage to solve the HappyCat function with a given accuracy of 10e-8.

14. HGBat Function

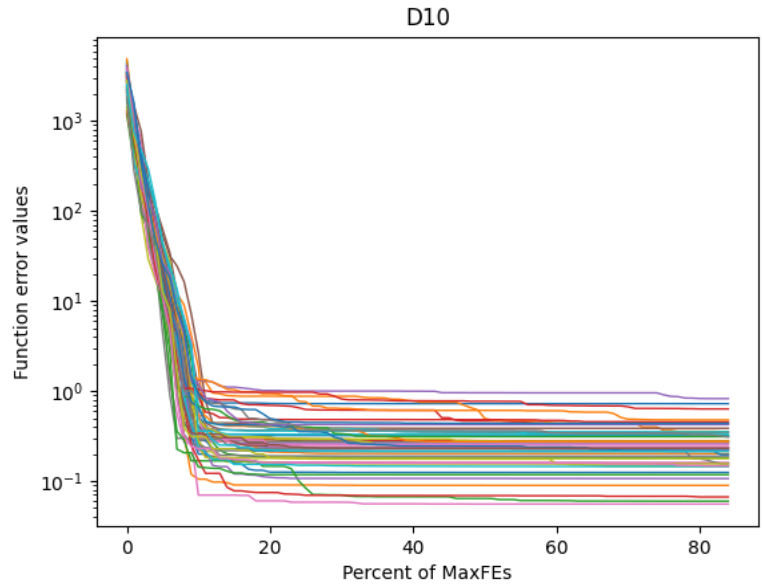

**Fig. 21.** Percentage of MaxFEs (for each of 51 runs) usage to solve the HGBat function with a given accuracy of 10e-8.

*C.  Binary tasks*

Below are the results of optimization of binary problems. Fig. 22-36 represent the change in error (the difference in absolute value between the optimum of the function and the current best solution of the algorithm) depending on the number of computational resources spent for each of the 51 runs of the GA. There were no restrictions on the maximum number of calculations of the objective function. Table 5 shows the minimum number of calculations required to achieve the optimum, maximum, mean, median and standard deviation among all (51) GA runs for each binary function for different dimensions.

**Table 5.** Solutions for D50, D100, D200, D500, D1000 binary optimization function.

| Function | Dimension | Min FEs | Max FEs | Mean FEs | Median FEs | Std FEs |
|---|---|---|---|---|---|---|
| **OneMax** | D50 | 190 | 775 | 395 | 397 | 106 |
| | D100 | 550 | 1450 | 894 | 874 | 223 |
| | D200 | 1378 | 3088 | 2090 | 2080 | 405 |
| | D500 | 4114 | 9775 | 5990 | 5734 | 1233 |
| | D1000 | 10927 | 30142 | 14960 | 14392 | 3067 |
| **Lead-ingOnes** | D50 | 1207 | 3529 | 2271 | 2269 | 535 |
| | D100 | 5464 | 12412 | 9163 | 9190 | 1675 |
| | D200 | 22348 | 44317 | 35287 | 35281 | 4599 |
| | D500 | 177787 | 278785 | 228227 | 233668 | 24999 |
| | D1000 | 962983 | 1329868 | 1140563 | 1138051 | 84765 |
| **Trap** | D50 | 190 | 505 | 308 | 298 | 68 |
| | D100 | 460 | 1009 | 725 | 739 | 120 |
| | D200 | 1144 | 2314 | 1628 | 1612 | 224 |
| | D500 | 4042 | 6545 | 5055 | 5077 | 595 |

1. OneMax Function

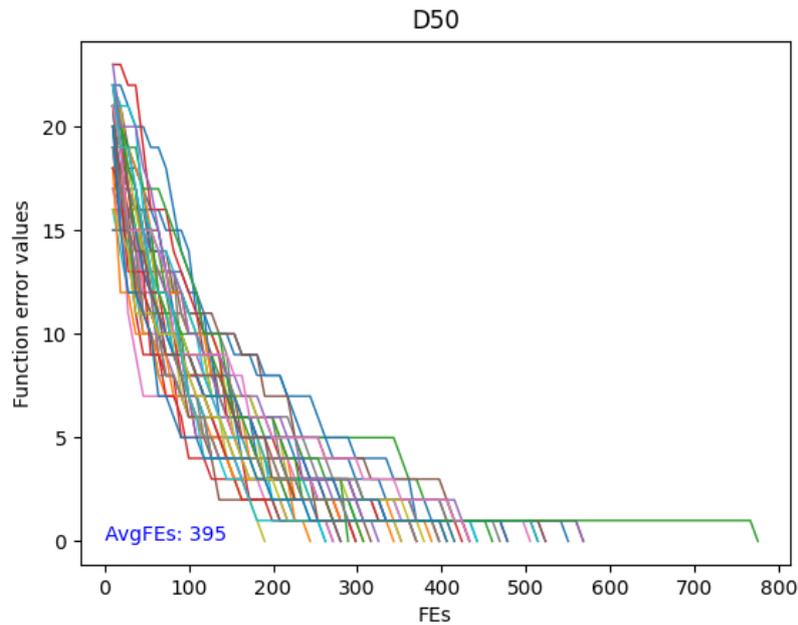

**Fig. 22.** The number of FEs required to achieve the optimum (for each of 51 runs) of OneMax function with a dimension 50.

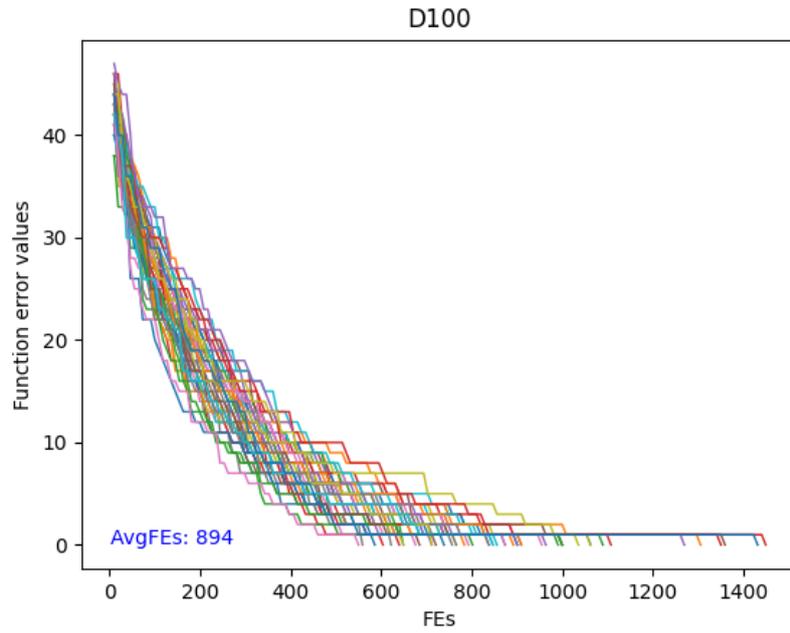

**Fig. 23.** The number of FEs required to achieve the optimum (for each of 51 runs) of OneMax function with a dimension 100.

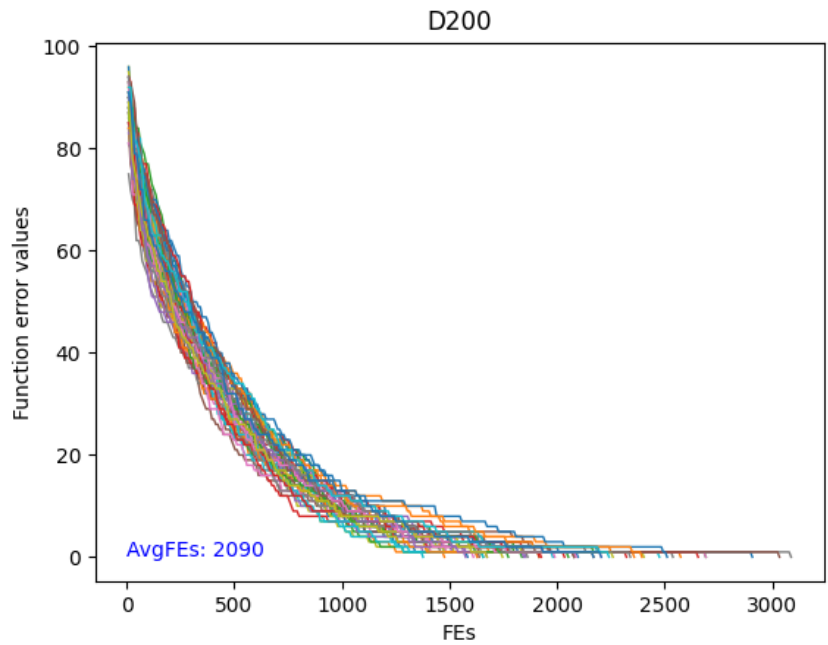

**Fig. 24.** The number of FEs required to achieve the optimum (for each of 51 runs) of OneMax function with a dimension 200.

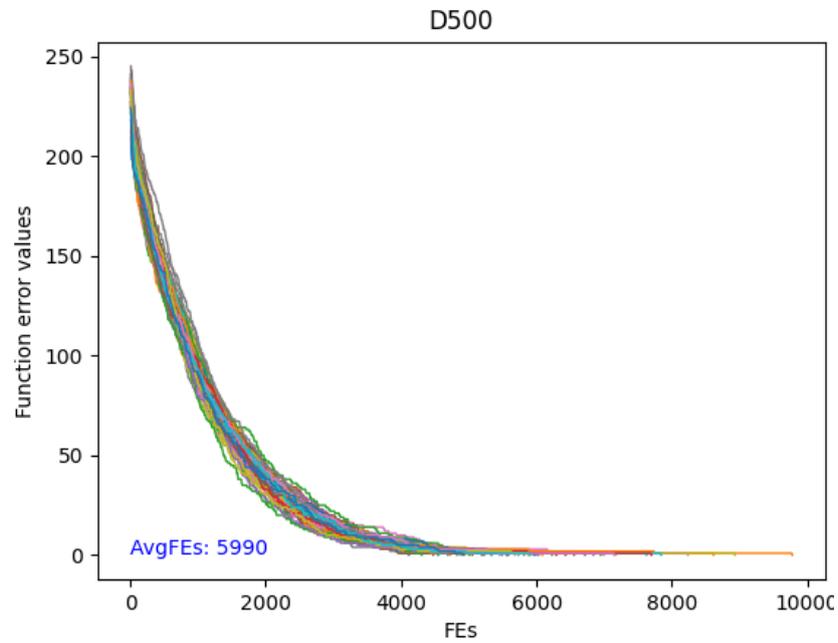

**Fig. 25.** The number of FEs required to achieve the optimum (for each of 51 runs) of OneMax function with a dimension 500.

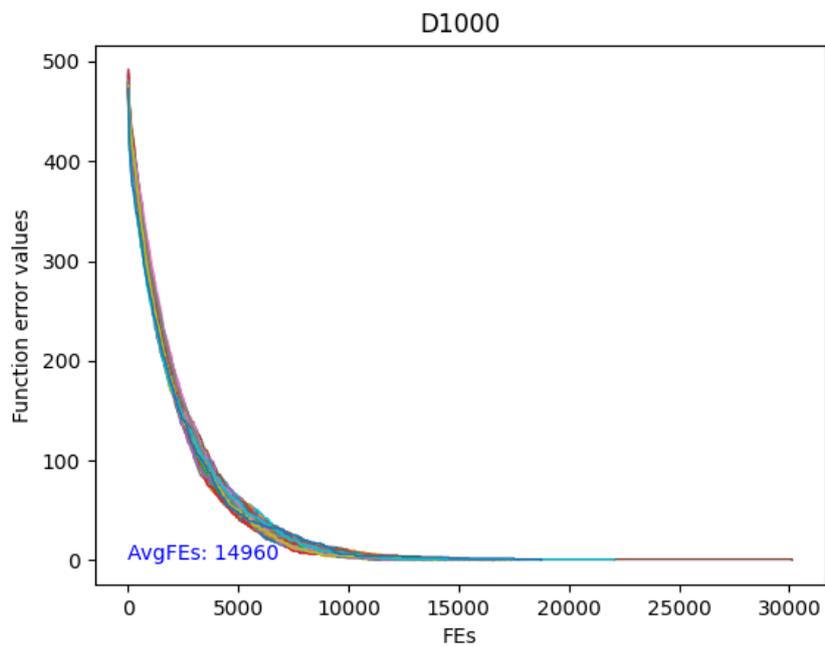

**Fig. 26.** The number of FEs required to achieve the optimum (for each of 51 runs) of OneMax function with a dimension 1000.

2. LeadingOnes Function

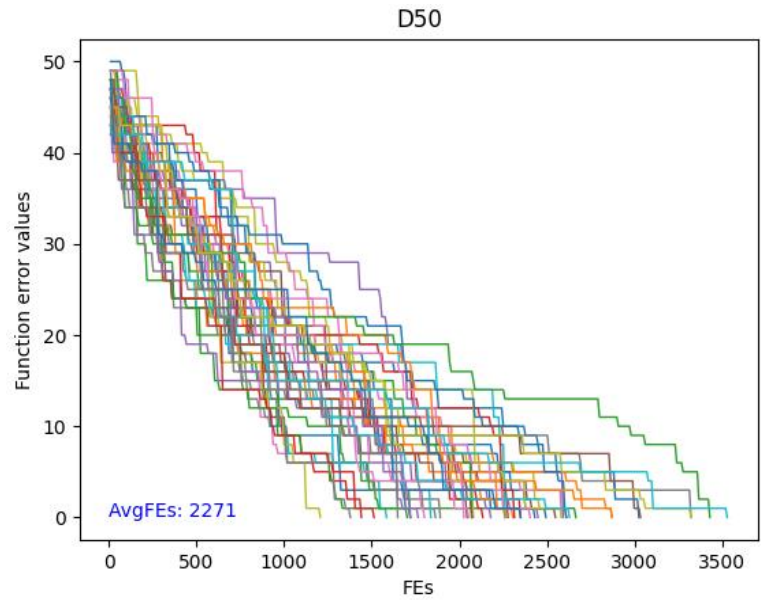

**Fig. 27.** The number of FEs required to achieve the optimum (for each of 51 runs) of LeadingOnes function with a dimension 50.

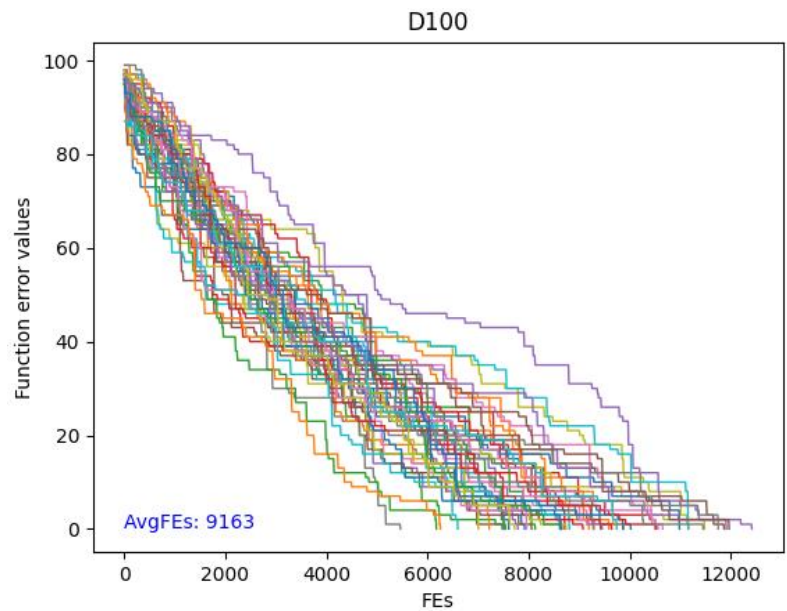

**Fig. 28.** The number of FEs required to achieve the optimum (for each of 51 runs) of LeadingOnes function with a dimension 100.

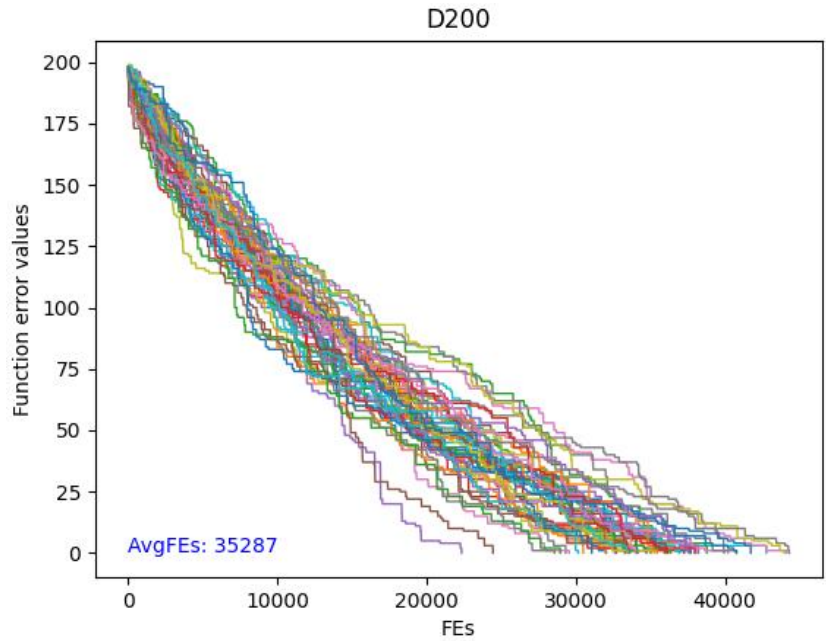

**Fig. 29.** The number of FEs required to achieve the optimum (for each of 51 runs) of LeadingOnes function with a dimension 200.

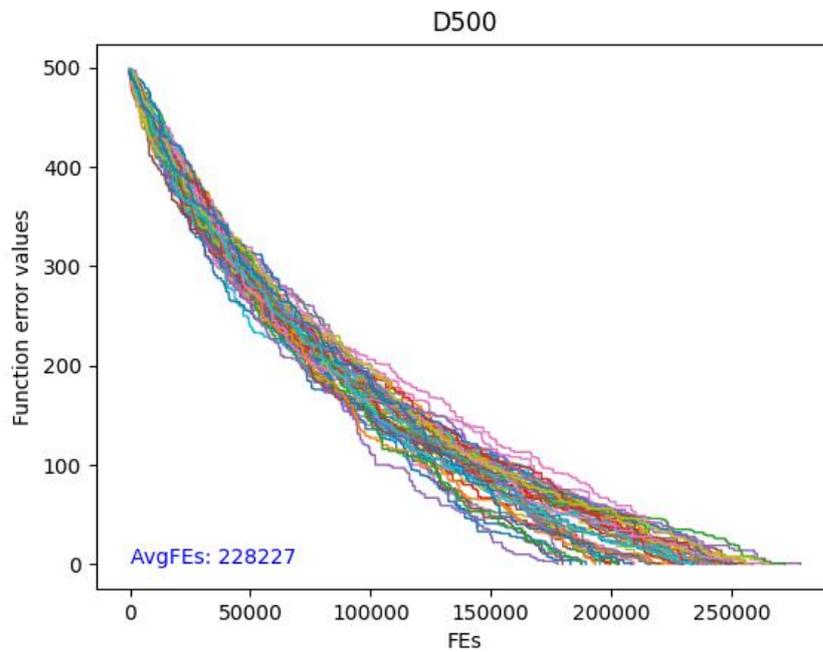

**Fig. 30.** The number of FEs required to achieve the optimum (for each of 51 runs) of LeadingOnes function with a dimension 500.

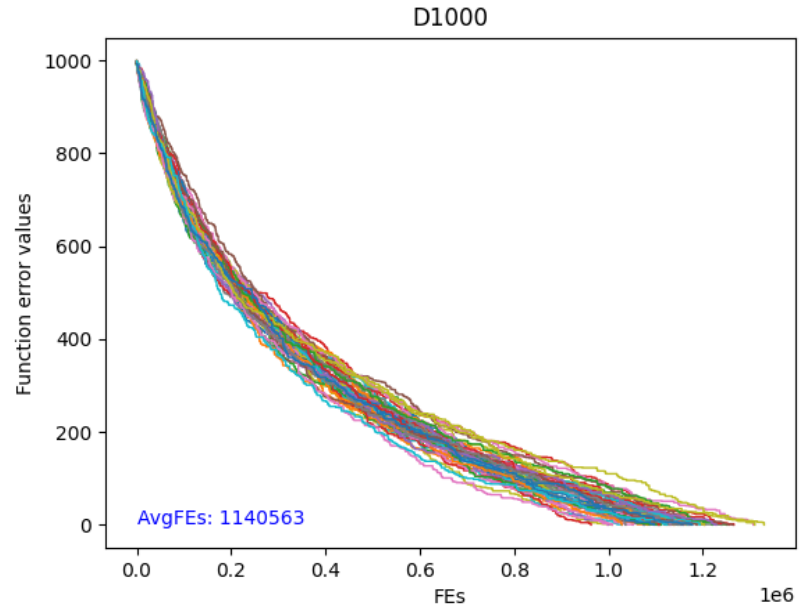

**Fig. 31.** The number of FEs required to achieve the optimum (for each of 51 runs) of LeadingOnes function with a dimension 1000.

3. Trap Function

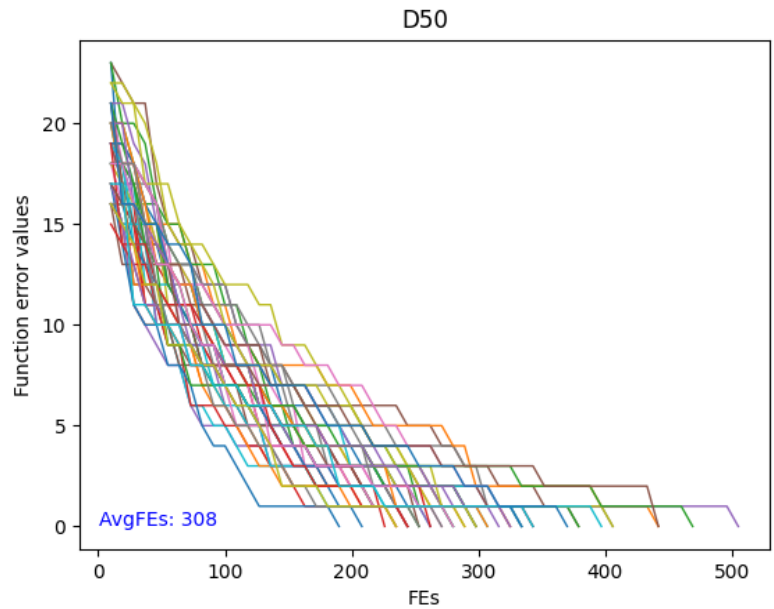

**Fig. 32.** The number of FEs required to achieve the optimum (for each of 51 runs) of Trap function with a dimension 50.

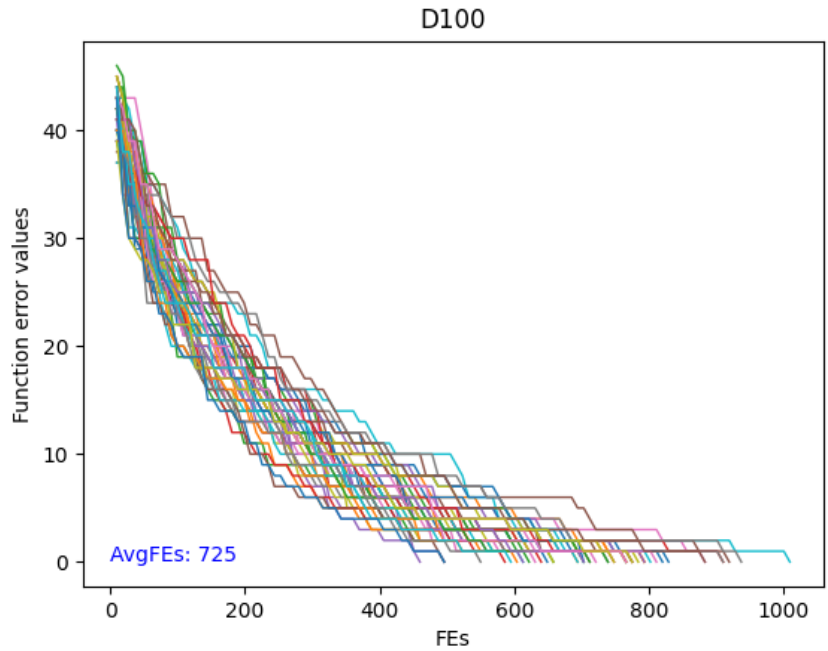

**Fig. 33.** The number of FEs required to achieve the optimum (for each of 51 runs) of Trap function with a dimension 100.

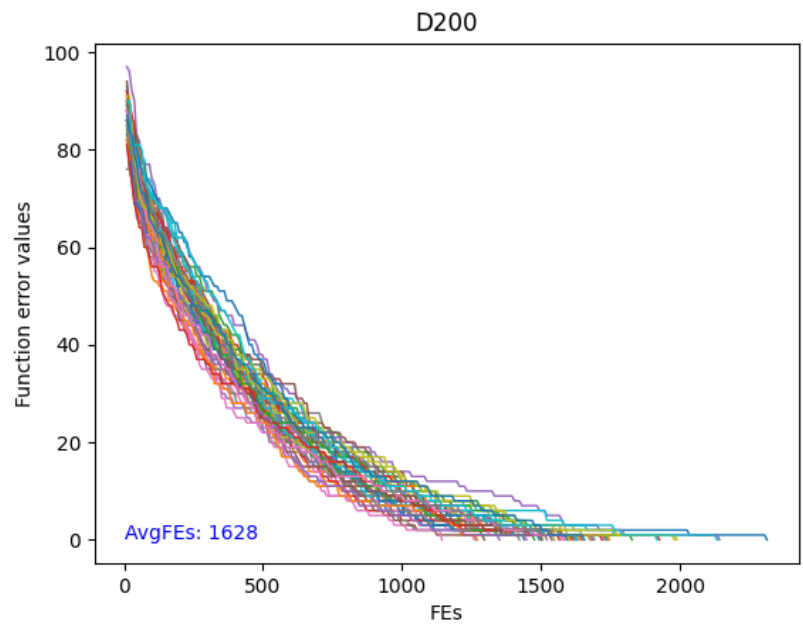

**Fig. 34.** The number of FEs required to achieve the optimum (for each of 51 runs) of Trap function with a dimension 200.

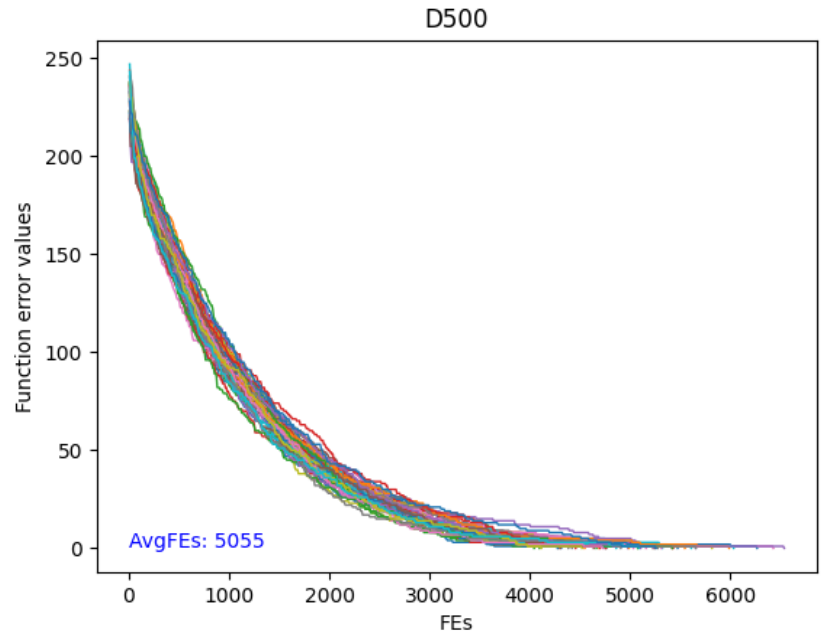

**Fig. 35.** The number of FEs required to achieve the optimum (for each of 51 runs) of Trap function with a dimension 500.

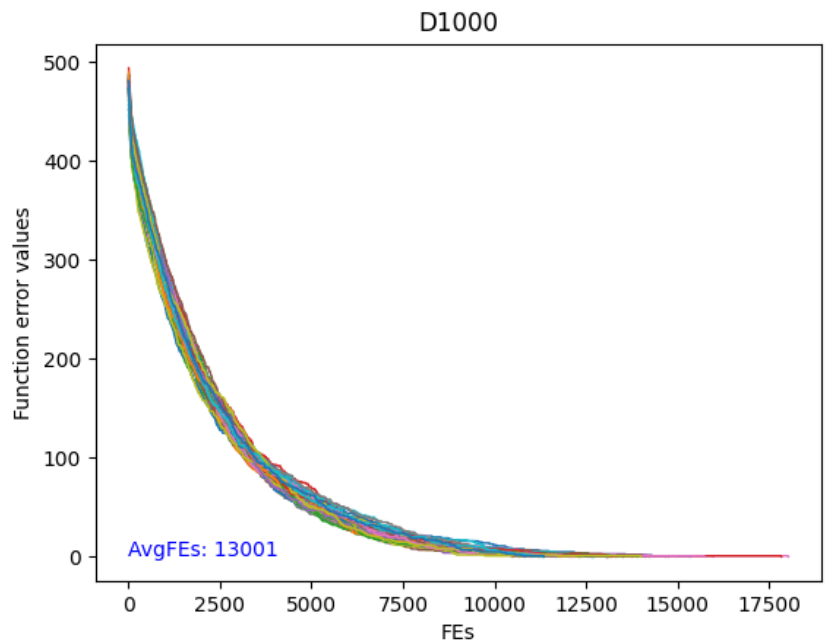

**Fig. 36.** The number of FEs required to achieve the optimum (for each of 51 runs) of Trap function with a dimension 1000.

# 4    Discussion

The results of testing the BaumEvA library showed excellent optimization results for binary tasks (Table 5), unimodal and multimodal functions (Table 4).

For some multimodal functions, the required accuracy is not achieved. Optimization Rosenbrock's function (Fig. 10) with a normal mutation, the required error value of 10e-8 is not achieved, while the best individual is in one of the last generations. Let's look at the behavior of the error when MaxFEs increases by 160 times, the Fig. 37 is shown below. The error continues to decrease. We can conclude that the required result is achievable, but only by going beyond the limitations on the number of calculations. Changes in population size, selection, crossing, and mutation do not make a significant contribution.

Problems were also discovered with the execution time of the genetic algorithm on multimodal problems with high dimensions, D50 and higher. This may be due both to the calculations of the objective function and to the implementation of the algorithm itself.

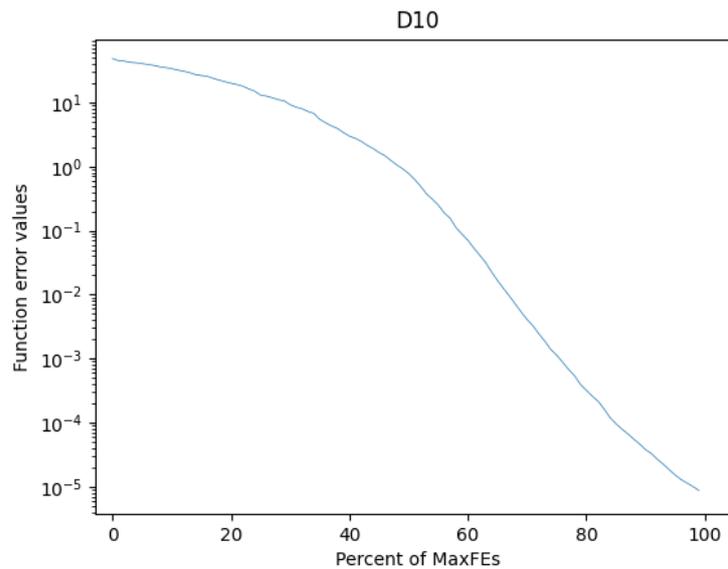

**Fig. 37.** Function error value of the Rosenbrock's function (single run) with MaxFEs increased by 160 times.

# 5    Conclusions

At the conclusion of the report on testing the BaumEvA library of evolutionary algorithms, the following conclusions can be drawn.

The library provides an efficient implementation of the genetic algorithm, in particular the binary genetic algorithm and with Gray codes.

Testing has shown the high efficiency of the algorithms in solving problems of unconditional optimization of Boolean functions, unimodal continuous functions and conditional optimization. Problems may arise in optimizing some multimodal functions, for example the Rastrigin's functions, see Fig. 11.

The library is highly scalable and can be used to solve problems of varying complexity and data volume, but sometimes with large dimensions and very small steps (10e-8) calculations can take a long time.

Also, in some cases, the performance of the library may depend on the quality of the settings of algorithms and parameters, which requires certain knowledge and experience from the user.

Overall, the evolutionary algorithms library is a powerful tool for solving a variety of optimization problems. It offers a wide range of capabilities and flexible configuration of algorithms, which allows it to be effectively used in various fields of science and technology. Further development of this library can be aimed at improving performance, expanding functionality and supporting parallel computing.